\documentclass[10pt,twocolumn,letterpaper]{article}

\usepackage[pagenumbers]{wacv} 

\usepackage{graphicx}
\usepackage{amsmath}
\usepackage{amssymb}
\usepackage{booktabs}
\usepackage{multirow, multicol}
\usepackage{bbm}
\usepackage{algorithm}
\usepackage{algpseudocode}
\usepackage{stfloats}
\usepackage[table, dvipsnames]{xcolor}

\usepackage[pagebackref,breaklinks,colorlinks]{hyperref}

\usepackage[capitalize]{cleveref}
\crefname{section}{Sec.}{Secs.}
\Crefname{section}{Section}{Sections}
\Crefname{table}{Table}{Tables}
\crefname{table}{Tab.}{Tabs.}

\def\wacvPaperID{1597} 
\def\confName{WACV}
\def\confYear{2025}

\begin{document}

\title{OT-VP: Optimal Transport-guided Visual Prompting for Test-Time Adaptation}

\author{Yunbei Zhang\textsuperscript{1}, Akshay Mehra\textsuperscript{1}, Jihun Hamm\textsuperscript{1}\\
{\small \textsuperscript{1}Tulane University}\\ 
{\tt\small\{yzhang111, amehra, jhamm3\}@tulane.edu}\\
}

\maketitle
\begin{abstract}
    Vision Transformers (ViTs) have demonstrated remarkable capabilities in learning representations, but their performance is compromised when applied to unseen domains. Previous methods either engage in prompt learning during the training phase or modify model parameters at test time through entropy minimization. The former often overlooks unlabeled target data, while the latter doesn't fully address domain shifts. In this work, our approach, Optimal Transport-guided Test-Time Visual Prompting (OT-VP), handles these problems by leveraging prompt learning at test time to align the target and source domains without accessing the training process or altering pre-trained model parameters. This method involves learning a universal visual prompt for the target domain by optimizing the Optimal Transport distance.OT-VP, with only four learned prompt tokens, exceeds state-of-the-art performance across three stylistic datasets—PACS, VLCS, OfficeHome, and one corrupted dataset ImageNet-C.  Additionally, OT-VP operates efficiently, both in terms of memory and computation, and is adaptable for extension to online settings. The code is available at \href{https://github.com/zybeich/OT-VP}{https://github.com/zybeich/OT-VP}.
    
\end{abstract} 
\section{Introduction}
\label{sec:intro}

The remarkable successes of Deep Neural Networks (DNNs) are often tempered by the challenges posed by discrepancies between training and testing data distributions \cite{recht2019imagenet, hendrycks2019benchmarking, koh2021wilds}.
Such discrepancies are not uncommon in real-world applications, where variations in data due to natural differences and stylistic changes can significantly impact model performance \cite{ li2017deeper}.
Though Domain Generalization (DG) has been proposed as a solution to ensure model robustness across unseen domains \cite{blanchard2011generalizing, zhou2022domain}, fully achieving this remains a challenge. 
To address this limitation, a new research direction has emerged, concentrating on enhancing model performance directly at test time \cite{wang2021tent, NEURIPS2021_1415fe9f}.  
This approach allows models to leverage unlabeled test data from target domains. 
This data offers insights into the target distribution, insights that are typically inaccessible in the DG framework. 
Test-time adaptation, as demonstrated in \cite{NEURIPS2021_1415fe9f}, surpasses the capabilities of many existing DG strategies by utilizing immediate, real-world data to refine model accuracy.
Inspired by these insights, our work pivots towards exploring test-time adaptation (TTA) as a strategic response to the challenges of domain shifts, aiming to harness the full potential of DNNs in unseen environments.

\begin{figure}[t]
    \centering
    \includegraphics[width=1\linewidth]{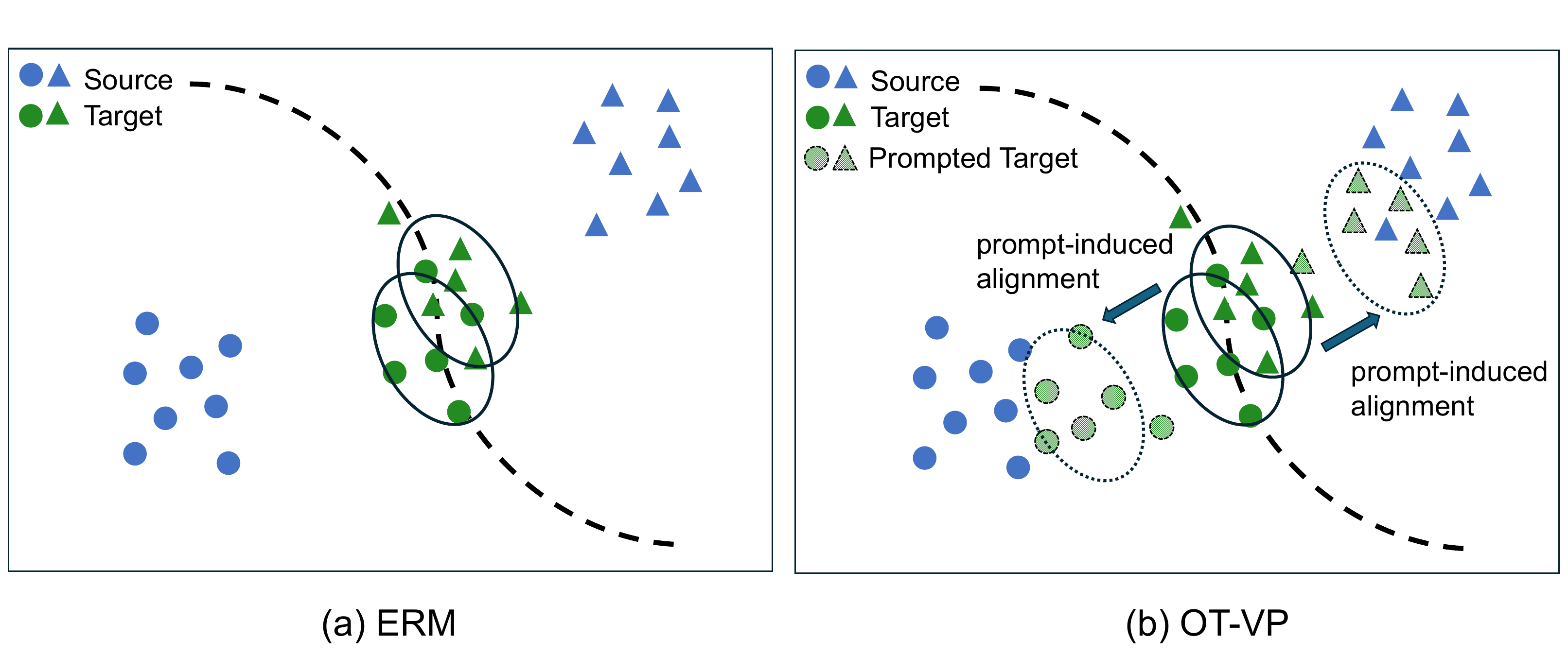}
    \caption{Motivation of our approach. (a) An ERM model trained on the source domain struggles to adapt to the target domain due to domain shifts. (b) Our method (OT-VP) optimizes a visual prompt by minimizing the Optimal Transport distance to align the target distribution (indicated as an ellipse) with the source distribution without changing the decision boundary.
    }
    \label{fig:motivation}
    \vspace{-0.5cm}
\end{figure}
\begin{figure*}[t]
    \centering
    \includegraphics[width=0.9\linewidth]{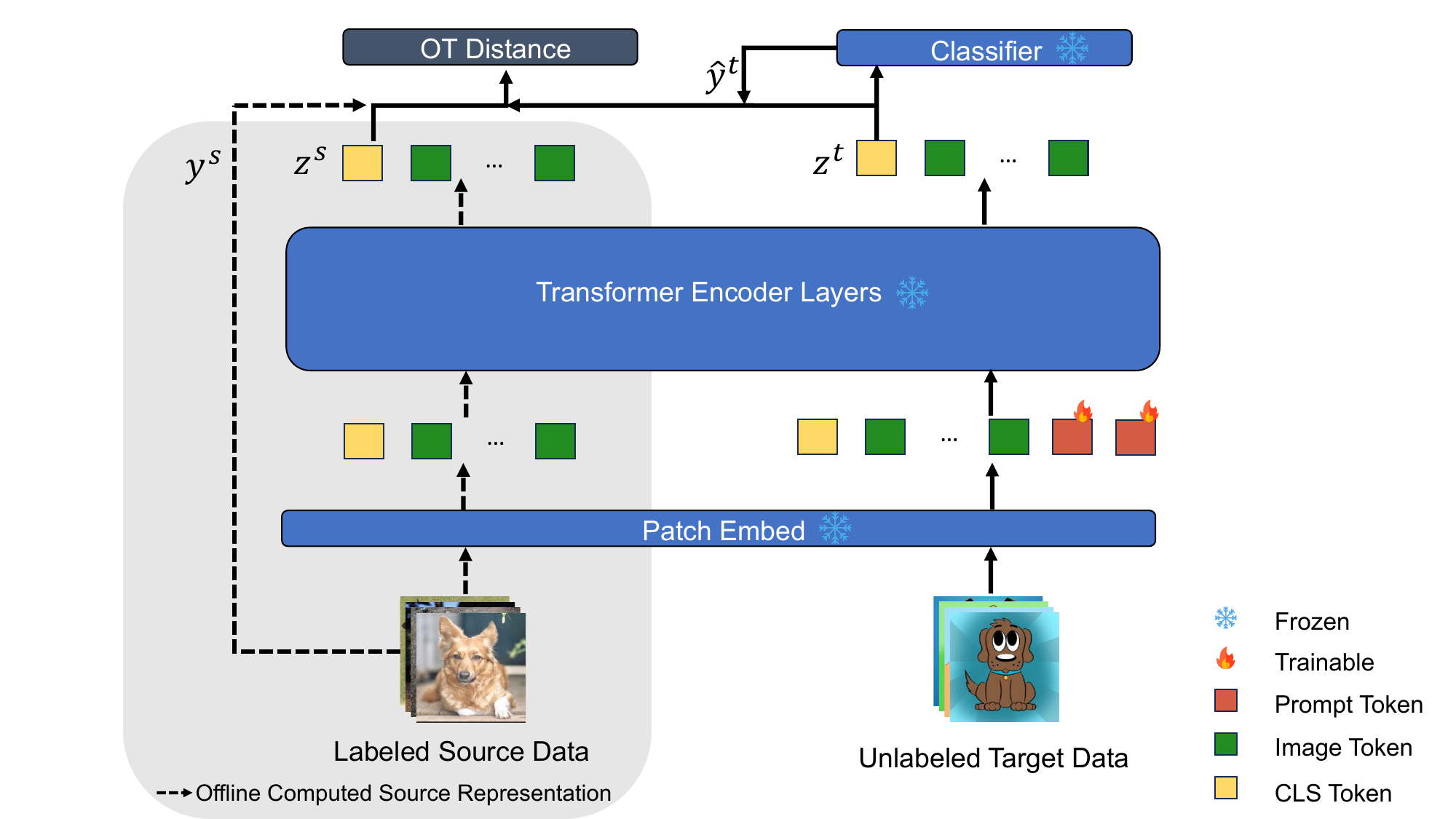}
    \caption{An overview of our proposed OT-VP method. At test time, unlabeled target data are processed through a frozen pre-trained ViT model with only the prompt tokens (indicated in red) being trainable.
    This generates target representations ($z^t$) and pseudo-labels ($\hat y^t$). 
    We then align these with actual source labels ($y^s$) and offline-computed (the grey shadowed area) source representations ($z^s$) via Optimal Transport (OT) distance. 
    The visual prompts are iteratively optimized based on this distance to align the source and target domain data more closely. 
    }
    \label{fig:workflow}
    \vspace{-0.5cm}
\end{figure*}
Our approach is underpinned by leveraging visual prompt learning, designed to seamlessly bridge the gap between source and target domains during test time. 
Vision Transformers (ViTs), known for their remarkable achievements across a spectrum of computer vision tasks, serve as the backbone of our approach \cite{dosovitskiy2021an, liu2021swin}. 
The self-attention mechanism within ViTs enables comprehensive modeling of relationships between various segments of an image, making them an ideal foundation for our work \cite{zheng2022prompt, naseer2021intriguing, raghu2021vision}.
Visual prompt learning emerges as a prominent strategy for fine-tuning ViTs without altering model parameters on specialized downstream tasks \cite{jia2022visual}.
This technique involves embedding task-specific knowledge directly into the input tokens, allowing for interaction with prompt tokens via self-attention layers. 
Such interactions enable the network to grasp the essence of the task, provided the prompt contains ample informative content. 
Although visual prompt tuning is lauded for its precision in task-oriented learning, it conventionally depends on the availability of labeled data for the creation of impactful prompt tokens—a requirement not met in the TTA context.

Current approaches in prompt learning typically incorporate prompts during the training phase, leveraging source data to create representative prompts \cite{zheng2022prompt}, or tuning prompts based on labeled source data at test time \cite{gao2023visual} for application in target domains. 
However, these conventional strategies often do not directly address the distribution shifts observed in the target domain, shown in Fig. \ref{fig:motivation} (a).  
\cite{zheng2022prompt} generates input-specific prompts based on a limited set of augmented target data, potentially missing the broader differences between source and target domains. Adopting a universal prompt for the target domain leverages more data and reduces computation time. Yet, simply learning this universal prompt by minimizing entropy doesn't effectively tackle domain shifts or enhance performance, as evidenced by our experiments in Sec. \ref{sec:ablation_study}.

To enable ViT models to adapt effectively at test time, it's imperative to address the fundamental challenge of reducing the distribution gap between source and target domains. 
As shown in Fig. \ref{fig:motivation}, this step is crucial for enhancing the model's ability to generalize to new, unseen data. 
Many metrics, such as Maximum Mean Discrepancy (MMD), Kullback-Leibler (KL) divergence, and Optimal Transport (OT) distance, have been explored to address distribution shifts \cite{NIPS2006_b1b0432c, gretton2012kernel, pmlr-v139-le21a, mansour2009domain}. 
Among these, Optimal Transport (OT) is distinguished as particularly effective due to its ability to utilize the geometry of the underlying space \cite{kantorovich1942translocation, peyre2019computational}. 
Numerous studies have demonstrated a strong correlation between OT distance and target performance, supported by thorough theoretical analyses \cite{shen2018wasserstein, mehra2023analysis, tan2021otce}. 
Inspired by these insights, we propose Optimal Transport-guided Test-Time Visual Prompting (OT-VP), a test-time distribution alignment method using prompt learning. 
Specifically, OT-VP explicitly minimizes the Optimal Transport distance between the source representation, computed offline, and the target representation without accessing the training process and modifying the pre-trained model parameters. 
As illustrated in Fig. \ref{fig:workflow}, for a given target dataset, we pass the unlabeled target image through the visual encoder with learnable prompts to get the target representation and pseudo-labels. 
Then we update the prompts by minimizing the OT distance.
This process is repeated iteratively to ensure that the prompts are well-aligned with the target distribution.

We comprehensively evaluate the efficacy of OT-VP across three stylistic datasets: PACS, VLCS, and OfficeHome, as well as one corrupted benchmark: ImageNet-C.
Beyond the conventional multi-source domain generalization (DG) setting, which typically leaves one domain out as the target while treating the remainder as sources, we also explore a single-source setting for stylistic datasets. This latter setting, often overlooked in existing literature \cite{NEURIPS2021_1415fe9f, zheng2022prompt}, involves using one domain as the source and a different one as the target. Training on a smaller dataset makes adaptation to the new target domain more challenging, thereby providing a rigorous test of the model's generalization ability. Based on the ViT architecture, OT-VP achieves stable and significant improvements for the source pre-trained model, enhancing average accuracy across all three stylistic datasets by $5.0\%$ in the single-source setting and $1.5\%$ in the multi-source setting, respectively. Moreover, OT-VP shows an 11.5\% improvement on the corrupted dataset. Notably, OT-VP surpasses state-of-the-art (SOTA) algorithms on both stylistic and corrupted datasets with fewer trainable parameters and less computational demand, underscoring its effectiveness and adaptability. Additionally, our method can be easily applied in more complex online settings.
Our contribution can be summarized as follows:

\begin{enumerate}
    \item[\textbullet] We propose OT-VP, a novel test-time adaptation that reduces the domain gap between source and target through minimizing Optimal Transport distance, facilitating prompts to better adapt to unlabelled target tasks.

    \item[\textbullet] Unlike existing methods, OT-VP does not require access to the training process or modifications to the pre-trained model's parameters. Instead, OT-VP merely injects a few prompt tokens into the input layer, enhancing memory and computation efficiency and preventing catastrophic forgetting.
    
    \item[\textbullet] We demonstrate through extensive experimentation that OT-VP consistently improves the performance of pre-trained models across a variety of settings, outperforming existing SOTA methods.

\end{enumerate}
\section{Related Work}
\label{sec:related_work}
\noindent\textbf{Prompting for Vision Transformers.} 
Vision Transformers (ViTs) have achieved state-of-the-art results in image classification \cite{dosovitskiy2021an,liu2021swin}, yet adapting ViTs to unseen target domains without labels is still a challenge. 
Many recent methods enhance ViTs' transferability by learning visual prompts as continuous, learnable vectors trained in an end-to-end manner with frozen model parameters \cite{jia2022visual, bahng2022visual, wang2022learning}. 
These approaches, however, typically rely on labeled target data for prompt training. For instance, \cite{zheng2022prompt} introduces DoPrompt, a Domain Generalization (DG) method that learns visual prompts for source domains and generates input-specific prompts at test time, though it requires alterations to the training process, limiting its practical utility. Similarly, DePT \cite{gao2023visual} tunes prompts using labeled source data before adapting them to the target domain with a memory bank and a teacher-student system.
In contrast, our approach learns a universal prompt for the target domain, effectively minimizing the distribution shift between source and target domains without requiring access to the training setup or any preliminary prompt tuning.

\vspace{12pt}
\noindent\textbf{Test-Time Adaptation (TTA).} TTA \cite{sun2020test, wang2021tent, liang2023ttasurvey, liang2020we, liang2021source, prabhudesai2023diffusion, mummadi2021test,  NEURIPS2021_1415fe9f, mirza2022norm, schneider2020improving, mirza2023actmad, gao2023visual, niu2024test, niu2023towards, mehra2024fly} aims to enhance the performance of pre-trained models when deployed in unseen domains. Traditionally, the majority of studies have focused on self-training and entropy regularization to align the model with the target data distribution. Notably, most existing TTA approaches have been developed for Convolutional Neural Networks (CNNs), and these methodologies are not directly applicable to Vision Transformers (ViTs) due to significant architectural differences, such as the absence of batch normalization layers in ViTs \cite{NEURIPS2021_1415fe9f}. This disparity highlights the need for devising TTA strategies specifically tailored to the unique architecture of ViTs, a gap our work seeks to fill.

Furthermore, contemporary methods predominantly employ entropy minimization or consistency maximization often neglecting to directly confront domain shifts \cite{wang2021tent, gao2023visual}. These approaches typically involve modifications to the source pre-trained model's parameters, which can lead to a risk of catastrophic forgetting \cite{kirkpatrick2017overcoming, gan2023decorate}. In contrast, our method uniquely leverages visual prompts while maintaining the entire pre-trained model frozen, thus effectively bridging the gap between source and target domains without altering underlying model parameters.

\begin{table*}[ht]
\centering
\resizebox{1.0\textwidth}{!}{%
\begin{tabular}{cccccccccccccccc|cc}
\toprule
 \textbf{Algo.} & \textbf{Gauss.} & \textbf{Shot} & \textbf{Impul.} & \textbf{Defoc.} & \textbf{Glass} & \textbf{Motion} & \textbf{Zoom} & \textbf{Snow} & \textbf{Frost} & \textbf{Fog} & \textbf{Brit.} & \textbf{Contr.} & \textbf{Elastic} & \textbf{Pixel.} & \textbf{JPEG} & \textbf{Avg.} & \textbf{Gain}\\
\midrule
\textbf{ERM} & 56.8 & 56.8 & 57.5 & 46.9 & 35.6 & 53.1 & 44.8 & 62.2 & 62.5 & 65.7 & 77.7 & 32.6 & 46.0 & 67.0 & 67.6 & 55.5 & 0.0 \\

\textbf{Tent-C} & 54.7 & 55.6 & 55.3 & 44.2 & 34.1 & 51.7 & 42.6 & 61.9 & 60.3 & 63.5 & 74.7 & 31.3 & 45.7 & 66.1 & 63.2 & 53.7 & -1.8 \\

\textbf{Tent-BN} & 57.5 & 57.6 & 58.2 & 50.0 & 38.1 & 54.9 & 47.4 & 64.7 & 65.1 & 71.3 & 78.0 & 59.1 & 50.0 & 67.7 & 68.9 & 59.2 & +3.7  \\

\textbf{Tent-LN} & 60.3 & 61.4 & 61.7 & 58.3 & 56.5 & 58.6 & 59.1 & 54.3 & 64.7 & 2.6 & 78.1 & 66.3 & 61.0 & 71.5 & 70.2 & 59.0& +3.5 \\

\textbf{T3A} & 56.4 & 56.9 & 57.3 & 47.9 & 37.8 & 54.3 & 46.9 & 63.6 & 60.8 & 68.5 & 78.1 & 38.3 & 50.0 & 67.6 & 69.1 & 56.9 & +1.4 \\

\textbf{DePT} & 54.4 & 58.2 & 58.3 & \textbf{59.4} & \textbf{58.0} & \underline{61.6} & \underline{59.8} & \textbf{70.6} & \underline{67.3} & \underline{73.5} & 79.2 & 64.5 & \underline{67.8} & \underline{73.6} & 70.5 & \underline{65.1} & +9.6  \\

\rowcolor{lightgray}
\textbf{OT-VP-B} & \underline{60.9} & \underline{63.2} & \textbf{64.0} & 56.8 & 51.7 & 59.1 & 56.0 & 68.1 & 63.4 & 73.3 & \textbf{80.0} & \textbf{68.3} & 62.9 & 72.7 & \textbf{72.8} & 64.9 & +9.4 \\

\rowcolor{lightgray}
\textbf{OT-VP}  & \textbf{61.2} & \textbf{62.4} & \underline{61.9} & \underline{58.4} & \underline{57.6} & \textbf{63.2} & \textbf{64.1} & \underline{70.5} & \textbf{70.4} & \textbf{74.3} & \underline{79.2} & \textbf{66.5} & \textbf{69.4} & \textbf{73.9} & \underline{71.7} & \textbf{67.0} & +11.5 \\

\midrule
\midrule

\textbf{DePT (online)} &53.7 & 55.7 & \underline{55.8} & \underline{58.2} & \textbf{56.0} & \textbf{61.8} & \underline{57.1} & \textbf{69.2} & 66.6 & \underline{72.2} & 76.3 & 63.2 & \underline{67.9} & \underline{71.8} & 68.2 &  63.6 & +8.1 \\

\rowcolor{lightgray}
\textbf{OT-VP-B (online)} & \underline{61.0} & \underline{61.5} & \underline{62.6} & 54.2 & 52.5 & 55.6 & 54.2 & 62.9 & \underline{66.8} & 67.5 & \underline{78.6} & \underline{64.5} & 56.3 & 68.7 & \underline{68.3} & 62.3 & +6.8 \\

\rowcolor{lightgray}
\textbf{OT-VP (online)} & \textbf{61.1} & \textbf{62.0} & \textbf{62.7} & \textbf{58.4} & \underline{54.7} & \underline{59.1} & \textbf{58.7} & \underline{67.8} & \textbf{69.1} & \textbf{72.8} & \textbf{79.3} & \textbf{66.6} & \textbf{68.1} & \textbf{73.0} & \textbf{71.8} & \textbf{65.7}  & +10.2 \\

\bottomrule

\end{tabular}
}
\caption{Results for ImageNet-C across 15 domains. OT-VP surpasses existing methods in both offline and online settings. The highest accuracy is highlighted in bold and the second highest is shown with an underline for clear distinction.}
\label{tab:imagenetc}
\end{table*}
\section{Method}
\label{sec:method}
In this section, we present OT-VP. We begin with a discussion on the problem setup of TTA in Section \ref{sec:problem_setup_preliminaries}. 
This is followed by introductions of Vision Transformers in Section \ref{sec:vit} and Optimal Transport in Section \ref{sec:ot}. 
Finally, we describe our method in Section \ref{sec:ot-vp}, with the method's workflow illustrated in Figure \ref{fig:workflow}.

\subsection{Preliminaries}
\label{sec:problem_setup_preliminaries}
\textbf{Problem Definitions.} Denote the data from the source (target) domains as $\mathcal{D}^s=\{\mathbf{x}_i^s, y_i^s\}_{i=1}^{n_s}$ ($\mathcal{D}^t=\{\mathbf{x}_i^t, y_i^t\}_{i=1}^{n_t}$), where $\mathbf{x} \in \mathcal{X}$ represents the input image and $y\in \mathcal{Y}$ is the label.
The dataset $\mathcal{D}^s$ ($\mathcal{D}^t$) comprises samples that are identically and independently distributed (i.i.d.), characterized by some probability distribution $P^s(X, Y)$ ($P^t(X, Y)$).

In the context of TTA, the model $f$ is initially trained on the source domain, e.g. minimizing the empirical risk, 
\begin{equation}
\label{eq:erm}
    \mathop{\arg \min}\limits_{f} \frac{1}{n_s} \sum_{i=1}^{n_s} \ell(f(\mathbf{x}_i^s), y^s)
\end{equation}
where $\ell$ is a loss function. Throughout this paper, we refer to optimizing the model with Eq. \ref{eq:erm} as ERM. 
Generally, the model $f$ is structured as a composition $f = h \circ \phi$, with the feature extractor $\phi: \mathcal{X} \rightarrow \mathcal{Z}$ learning the input's representation, and the classifier $h: \mathcal{Z} \rightarrow \mathcal{Y}$ predicting the class label. 

For any unlabeled target domain $\mathcal{D}^t$, TTA aims to adapt model $f$ to bridge the performance gap under the assumption that the source domain and target domain share the same label set. In our approach, we employ a Vision Transformer as the model $f$, which remains fixed during adaptation.

\subsection{Vision Transformers}
\label{sec:vit}
A Vision Transformer (ViT) \cite{dosovitskiy2021an, liu2021swin} processes an input image $x$ by initially dividing it into $k$ patches $\{I_i\}_{i=1}^k$. 
An encoding layer $E$ is employed to transform the input patches into patch tokens, to which positional embedding are subsequently added to retain spatial information. 
The inputs to the transformer layers consist of these encoded patch tokens augmented with a special classification token \verb|[CLS]|. 
The ViT is composed of several sequential blocks, and each block contains an attention layer and a Multi-Layer Perceptron  (MLP) layer. 
The prediction of the vision transformer can be formulated as follows:
\begin{equation}
\begin{aligned}
    \texttt{[CLS]} &= \phi([\texttt{[CLS]}, E(I_1), ..., E(I_k)]), \\
    y &= h(\texttt{[CLS]}),
\end{aligned}
\end{equation}
where $[\cdot]$ represents concatenation of tokens. 

Incorporating a visual prompt into the ViT represents a parameter-efficient approach for fine-tuning or adapting the model, particularly when it is fixed \cite{jia2022visual, ge2023domain}. 
By introducing $l$ prompt tokens $\{\texttt{[Prompt]}_i\}_{i=1}^l=: \mathbf{\gamma}$, the prediction process can be reformulated as follows:
\begin{equation}
\begin{aligned}
    \texttt{[CLS]} &= \phi([\texttt{[CLS]}, \{E(I_i)\}_{i=1}^k, \gamma]) \\
    y &= h(\texttt{[CLS]})
\end{aligned}
\label{eq:infer_with_prompt}
\end{equation}
The optimal prompts can be optimized as follows when the labels are available:
\begin{equation}
\label{eq:supervised_vp}
    \mathbf{\gamma}^* = \mathop{\arg \min}\limits_{\mathbf{\gamma}} \mathbb{E}[\ell(f(\mathbf{x}; \mathbb{\gamma}), y)]
\end{equation}

\subsection{Optimal Transport}
\label{sec:ot}
Optimal Transport (OT) theory, tracing back to the Monge problem in 1781, evolved significantly with the introduction of the Kantorovich relaxation \cite{kantorovich1942translocation} in 1942. 
This advancement transformed OT into a robust framework for comparing distributions, shapes, and point clouds \cite{peyre2019computational}, leveraging the geometry of the underlying space. 
OT operates on a complete and separable metric space $\mathcal{X}$, utilizing continuous or discrete probability measures $P, Q \in \mathcal{P(X)}$. 
The Kantorovich formulation defines the OT problem as:
\begin{equation}
\mathrm{OT}_c(P, Q) := \mathop{\inf}\limits_{\pi \in \Pi(P, Q)} \int_{\mathcal{X} \times \mathcal{X}} c(\mathbf{x}_1, \mathbf{x}_2)d\pi (\mathbf{x}_1, \mathbf{x}_2),
\end{equation}
where $c(\cdot,\cdot): \mathcal{X}\times\mathcal{X} \rightarrow \mathbb{R^+}$ denotes a cost function, and $\Pi(P, Q)$ represents the set of all possible couplings or joint distributions over $\mathcal{X}\times\mathcal{X}$ with $P$ and $Q$ as their marginals. 
The term $W_p(P,Q) := \mathrm{OT}_c(P, Q)^{\frac{1}{p} }$ is referred to as the $p$-Wasserstein distance when the cost function $c(\mathbf{x}_1, \mathbf{x}_2)=d(\mathbf{x}_1, \mathbf{x}_2)^p$ for some $p\geq1$ where $d$ is a metric of $\mathcal{X}$.

In real-world applications, the true marginal distributions $P, Q$ are often unknown, leading to reliance on discrete empirical distributions $\hat{P}=\sum_{i=1}^{m}\mathbf{a}_i\delta_{\mathbf{x}_1^i}$ and $\hat{Q}=\sum_{i=1}^{n}\mathbf{b}_i\delta_{\mathbf{x}_2^i}$, with $\mathbf{a},\mathbf{b}$ as vectors in the probability simplex. 
The cost function then simplifies to an $m \times n$ cost matrix $\mathbf{C}$, where $\mathbf{C}_{ij} = c(\mathbf{x}_1^i, \mathbf{x}_2^j)$. 
For computational efficiency, the Sinkhorn algorithm \cite{cuturi2013sinkhorn} introduces an entropic regularizer to the OT problem, facilitating practical applications such as domain adaptation \cite{courty2016optimal} and the evaluation of distances between datasets \cite{alvarez2020geometric}. 
This regularized approach, which can be computed using \verb|POT| \cite{flamary2021pot}, allows for the computation of an optimal mapping from source to target domains.

\subsection{Test-time Adaptation with OT-VP}
\label{sec:ot-vp}
In the TTA setting, the absence of labeled target data presents a challenge for prompt optimization as traditionally conducted in Eq. \ref{eq:supervised_vp}. 
To address this, we introduce an unsupervised prompt adaptation strategy, termed \underline{\textbf{O}}ptimal \underline{\textbf{T}}ransport-guided Test-Time \underline{\textbf{V}}isual \underline{\textbf{P}}rompting (\textbf{OT-VP}). 
This method leverages unlabeled target dataset $\mathcal{D}^t$, passing it through the ViT encoder alongside learnable prompts to obtain a set of representations, as depicted in Figure \ref{fig:workflow}. 
Source representations, prepared in advance, are readily applied during test time to facilitate OT distance computation.

The essence of OT-VP lies in calculating the OT distance between the target and pre-computed source representations, using two distinct cost functions. Note that the source representations are pre-computed prior to adaptation, and the source data is not utilized during test time.
First, \textbf{OT-VP-B (base)} measures the cost between two representations devoid of label data, with the cost $c_0(\mathbf{z}^s, \mathbf{z}^t)$ defined as the Euclidean distance between source and target representations $\mathbf{z}^s:=h(\mathbf{x}^s)$ and $\mathbf{z}^t:=h(\mathbf{x}^s; \gamma)$:
\begin{equation}
\label{eq:ot_dist}
    c_0(\mathbf{z}^s, \mathbf{z}^t) = \|\mathbf{z}^s- \mathbf{z}^t||_2
\end{equation}

Second, \textbf{OT-VP} enriches this comparison by incorporating label or pseudo-label information, introducing a penalty term scaled by hyperparameter $\lambda$ for label mismatches between source and target data:
\begin{equation}
\label{eq:ot_dist_with_label}
    c_\lambda((\mathbf{z}^s, y^s), (\mathbf{z}^t, \hat y^t)) = \|\mathbf{z}^s- \mathbf{z}^t\|_2 + \lambda \cdot \mathbbm{1}_{\{y^s \neq \hat y^t\}}
\end{equation}
where $\hat y^t := f(\mathbf{x}^t; \mathbf{\gamma}))$ represents the pseudo label derived from the pre-trained model using the adaptively learned prompts.
Notably, when setting $\lambda$ to infinity in Eq. \ref{eq:ot_dist_with_label}, the OT distance is the Wasserstein Distance \cite{sinha2017certifying}, a well-established mathematical quantity that has proven effective for measuring distances between distributions.

The computed OT distance informs the prompt update process for the target dataset, optimizing the prompts to minimize the distance between the source and target distributions. 
This optimization is formalized as seeking the optimal prompts $\mathbf{\gamma}^*$ that minimize the OT cost, thereby aligning the target dataset's representation with that of the source:
\begin{equation}
\label{eq:ot-vp}
    \mathbf{\gamma}^* = \mathop{\arg \min}\limits_{\mathbb{\gamma}} \mathrm{OT}_c(P^s_\#, P^t_\#)
\end{equation}
where $P^s_\#$ is a joint distribution over source representations and source labels: $(\phi(\mathbf{x}^s), y^s)$, and $P^t_\#$ is a distribution over target representations and target pseudo labels: $(\phi(\mathbf{x}^t), \hat y^t)$.

During inference, we apply the optimized prompt tokens $\gamma^*$ to make predictions for a given target input $\mathbf{x}^t$, following the Eq. \ref{eq:infer_with_prompt}.


\begin{table}[ht]

\centering

\resizebox{0.5\textwidth}{!}{%

\begin{tabular}{lccc|cc}

\toprule

\textbf{Algo.} & \textbf{PACS} & \textbf{VLCS} & \textbf{OfficeHome} & \textbf{Avg.} & \textbf{Gain}   \\ 

\midrule

\textbf{ERM} & 64.5 & 63.8 & 66.7 & 65.0 & 0.0 \\

\textbf{DoPrompt} & 64.3 & 65.8 & 67.6 & 65.9 & +0.9 \\

\textbf{Tent-C} & 64.4 & 65.0 & 66.6 & 65.3 & +0.3 \\

\textbf{Tent-BN} & 69.0 & 58.5 & 67.6 & 65.0 & 0.0 \\

\textbf{T3A} & \underline{71.2} & \underline{67.7} & \underline{68.1} & \underline{69.0} & +4.0 \\

\textbf{DePT} & 67.1 & 66.2 & 67.1 & 66.8 & +1.8 \\

\midrule

\rowcolor{lightgray} 
\textbf{OT-VP-B} & 69.8 & 65.2 & 66.9 & 67.3 & +2.3 \\

\rowcolor{lightgray} 
\textbf{OT-VP} & \textbf{73.5} & \textbf{68.4} & \textbf{68.1} & \textbf{70.0} & +5.0 \\

\bottomrule

\end{tabular}

}

\caption{Accuracy ($\%$) in Single-Source Settings across Datasets and Algorithms. The last column represents the relative improvement over the baseline established by ERM on a ViT-Base model. Full results can be found in Appendix \ref{sec:full_results}.}

\label{tab:1src_avg}

\end{table}

\begin{table}[ht]

\centering

\resizebox{0.5\textwidth}{!}{%

\begin{tabular}{lccc|cc}

\toprule

\textbf{Algo.} & \textbf{PACS} & \textbf{VLCS} & \textbf{OfficeHome} & \textbf{Avg.} & \textbf{Gain}  \\ 

\midrule

\textbf{ERM} & 87.0 & 78.5 & 73.6 & 79.7 & 0.0 \\

\textbf{DoPrompt} & \underline{87.5}  & \underline{80.3} & 74.7 & \underline{80.8} & +1.1 \\

\textbf{Tent-C} & 87.2 & 78.8 & 73.6 & 79.9 & +0.2\\

\textbf{Tent-BN} & 87.2 & 78.5 & 74.6 & 80.1 & +0.4 \\

\textbf{T3A} & 87.4 & 80.0 & 74.7 & 80.7 & +1.0 \\

\textbf{DePT} & 87.3 & 80.1 & \underline{74.8} & 80.7 & + 1.0 \\
\midrule

\rowcolor{lightgray} 
\textbf{OT-VP-B} & 87.3 & 80.2 & 74.3 & 80.6 & +0.9 \\
\rowcolor{lightgray} 
\textbf{OT-VP} & \textbf{87.7} & \textbf{80.9} & \textbf{75.1} & \textbf{81.2} & +1.5 \\

\bottomrule

\end{tabular}

}

\caption{Accuracy ($\%$) in Multi-Source Settings across Datasets and Algorithms. The last column represents the relative improvement over the baseline established by ERM on a ViT-Base model. Full results can be found in Appendix \ref{sec:full_results}.}

\label{tab:3src_avg}

\end{table}

\section{Experiments}
\label{sec:experiments}

\subsection{Experimental Setup}

\noindent\textbf{Datasets.} We evaluate our method on four benchmark datasets for TTA. Three stylistic datasets: \textbf{PACS} \cite{li2017deeper}, \textbf{VLCS} \cite{fang2013unbiased}, \textbf{OfficeHome} \cite{Venkateswara_2017_CVPR}, and one corrupted dataset: \textbf{ImageNet-C} \cite{hendrycks2019benchmarking}. Details can be found in Appendix \ref{sec:dataset}

\vspace{12pt}
\noindent\textbf{Baselines.} We compare OT-VP against the following state-of-the-art TTA approaches: 

\begin{enumerate}
    \item Tent \cite{wang2021tent} is a TTA method that fine-tunes the batch normalization (BN) parameters to minimize prediction entropy during test time. 
    Notably, ViTs do not contain BN layers. Following the methodology outlined by \cite{NEURIPS2021_1415fe9f}, we adapt Tent for use with ViTs in two distinct manners. 
    Specifically, Tent-BN introduces a BN layer immediately preceding the linear classifier, allowing for the adjustment of normalization parameters within this BN layer. 
    Tent-C modulates the entire classifier to reduce prediction entropy. 
    Additionally, following \cite{niu2023towards}, Tent-LN directly adapts the layer normalization (LN) parameters.

    \item T3A \cite{NEURIPS2021_1415fe9f} begins with generating pseudo-prototype representations for each class using unlabeled target data and the pre-trained classifier. 
    Subsequently, the classification of each target sample is performed based on its distance to the pseudo-prototypes. 

    \item DoPrompt \cite{zheng2022prompt} learns domain-specific prompts for each source domain during the training phase. 
    Additionally, it employs a prompt adapter, a mechanism trained to craft an appropriate prompt for individual input images. 
    At test time, this prompt adapter generates an input-specific prompt based on the learned source domain prompts.
    Note that DoPrompt is NOT a TTA method as the prompts are optimized during the training time.

    \item DePT \cite{gao2023visual} first tunes prompts on the source labeled data and then adapts the prompts to the target domain, utilizing an additional memory bank and a teacher-student system under a hierarchical self-supervised objective. Besides the visual prompts, DePT also updates the classifier parameters. In contrast, our method simplifies the approach by adding prompts only to the first transformer layer, without requiring any auxiliary models. We keep the entire pre-trained model frozen, making only the prompt tokens trainable.
\end{enumerate}

\vspace{12pt}
\noindent\textbf{Implementation details.} 
For all experiments, we utilize the ViT-Base model pre-trained on ImageNet-1k, sourced from \texttt{timm} \cite{rw2019timm}. 
For implementation of our methods, we  use the \verb|DomainBed| library \cite{gulrajani2020search} \footnote{\href{https://github.com/facebookresearch/DomainBed}{https://github.com/facebookresearch/DomainBed}}. 
Adopting DoPrompt's strategy, we use 4 prompts, optimized via the AdamW optimizer \cite{loshchilov2017decoupled, kingma2014adam} with a 0.1 learning rate on the target validation set. 
The target dataset's batch size is 64, with prompt training capped at 100 and 50 training steps for ImageNet-C and the other three respectively to avoid overfitting. 
We set $\lambda$ in Eq. \ref{eq:ot_dist_with_label} to $10^4$. All experiments were conducted using an RTX 6000 Ada GPU.

\subsection{Experiments on PACS, VLCS, and OfficeHome}
Tables \ref{tab:1src_avg} and \ref{tab:3src_avg} summarize our experimental outcomes for PACS, VLCS and OfficeHome in single-source and multi-source settings, respectively. 
Each reported value is an average across three separate trials.
These trials vary by weight initialization and data splits. 
Comprehensive results can be found in the Appendix \ref{sec:full_results}.

\vspace{12pt}
\noindent\textbf{OT-VP consistently outperforms SOTA TTA methods across all datasets in both settings.}
Demonstrated by the results in Tables \ref{tab:1src_avg} and \ref{tab:3src_avg}, OT-VP remarkably boosts the ERM model's performance in both single-source and multi-source settings. 
Notably, OT-VP shows substantial improvements, with enhancements of $9.0\%$, $4.6\%$, and $1.4\%$ in the single-source scenario, and $0.7\%$, $2.4\%$, and $1.5\%$ in the multi-source scenario across the respective datasets: PACS, VLSC, and OfficeHome, highlighting its potent effect on model accuracy.

The contrast in performance between alternative approaches and our OT-VP is noteworthy. 
The observation that Tent-BN and Tent-C exhibit only nominal or no improvements, and sometimes even lead to decrease in performance on ViT models, aligns with the expectations considering the original design of Tent. 
Tent was initially conceived for models relying heavily on batch normalization layers for domain adaptation by adjusting these layers to minimize prediction entropy. 
However, ViTs operate on a different architectural principle, lacking batch normalization layers in their standard configuration. 
This fundamental discrepancy underscores the challenges of directly applying methodologies designed for CNNs to ViTs without modifications or considerations of architectural differences. 
Furthermore, T3A and DePT consistently improve performance over Tent in both experimental setups, though their gains are smaller than those achieved by OT-VP. 

\begin{figure*}
    \centering
    \begin{subfigure}{.5\textwidth}
      \centering
      \includegraphics[width=1\linewidth]{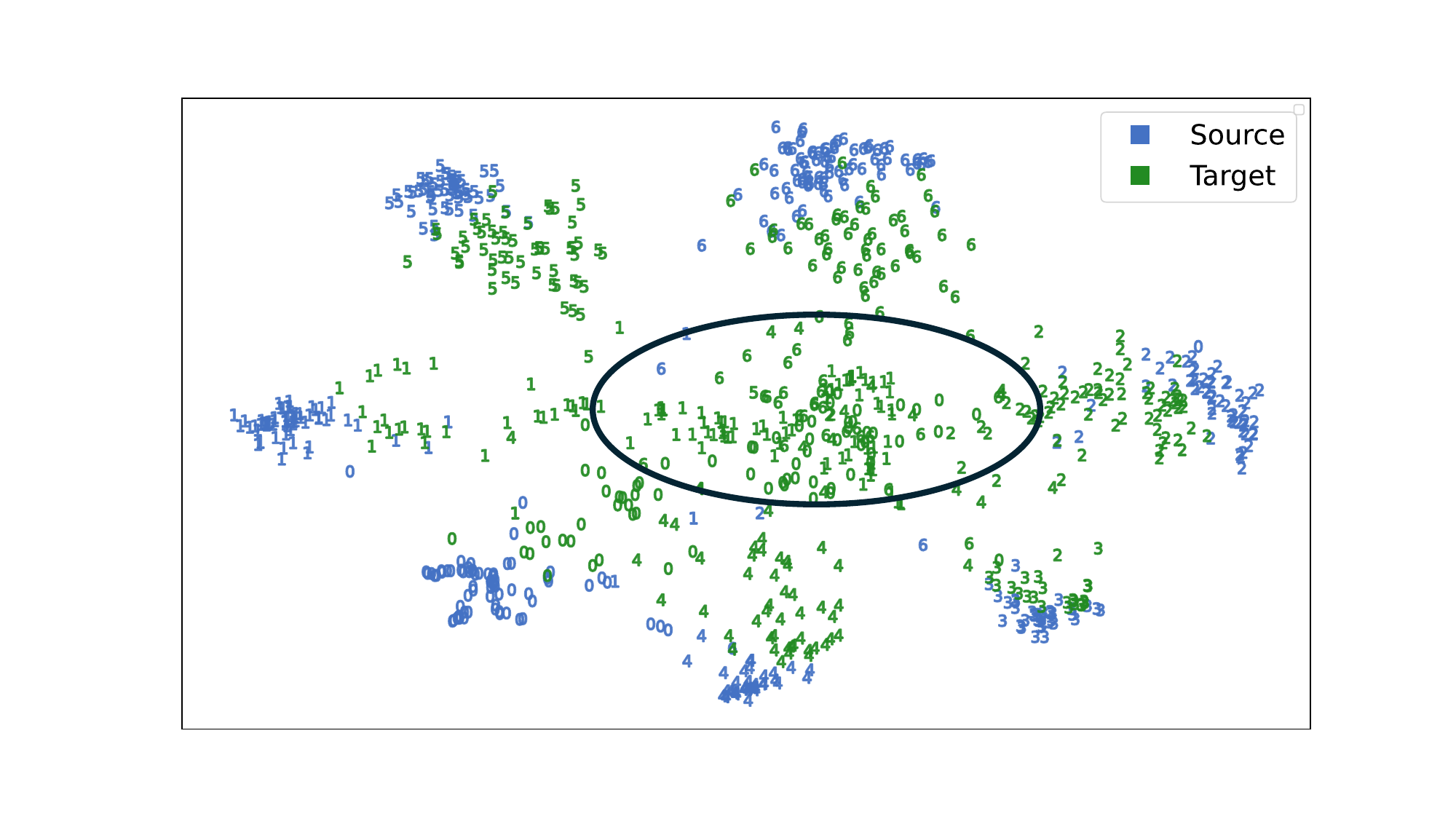}
      \caption{Before OT-VP}
      \label{fig:before_prompt}
    \end{subfigure}
    ~
    \begin{subfigure}{.5\textwidth}
      \centering
      \includegraphics[width=1\linewidth]{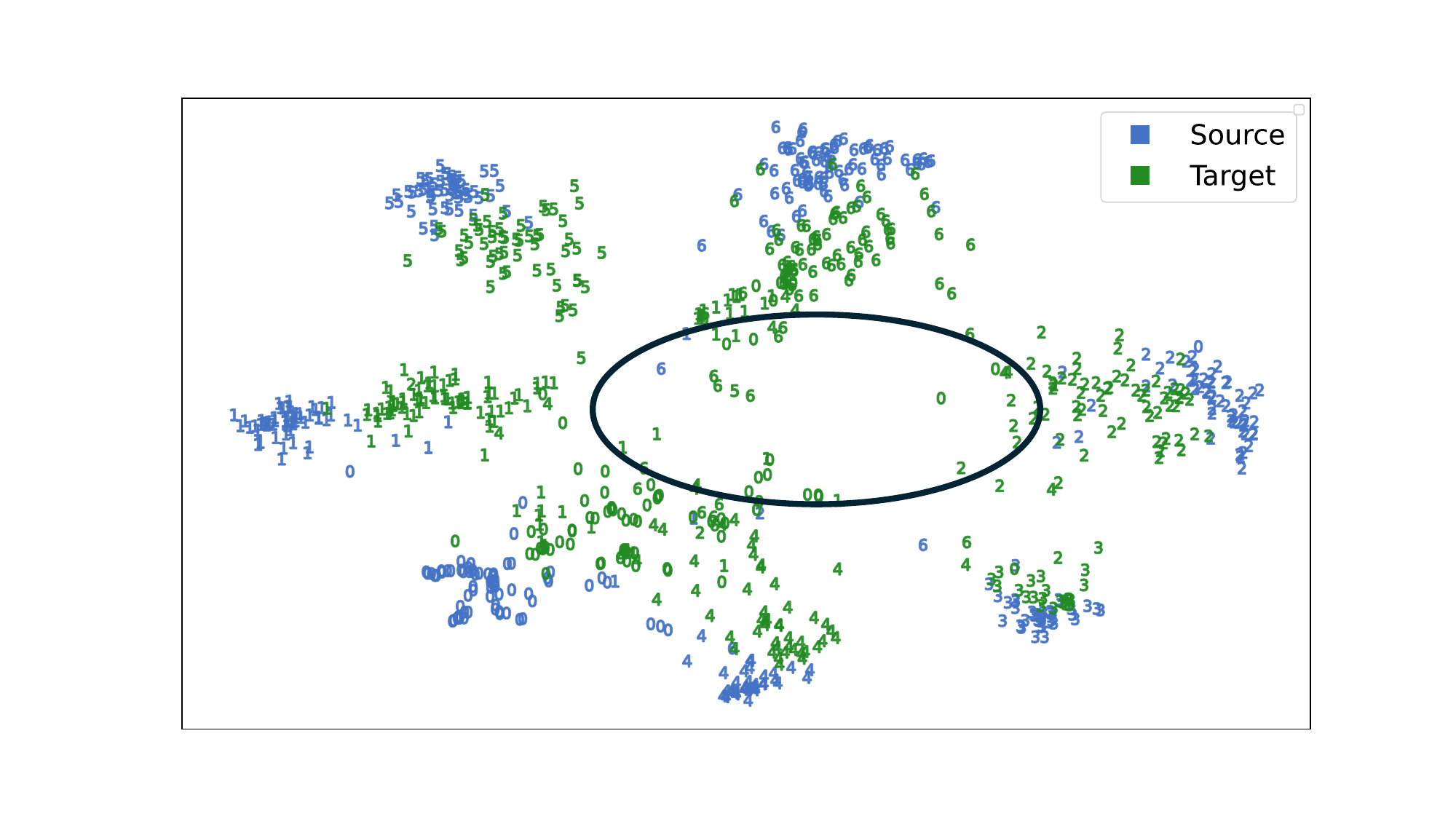}
      \caption{After OT-VP}
      \label{fig:after_prompt}
    \end{subfigure}
    \caption{{t-SNE visualization showcasing the impact of OT-VP.} The figures display the representation space before and after the application of OT-VP for A $\rightarrow$ C in the PACS dataset with the pre-trained ViT encoder. 
    Different numbers represent distinct class labels.
    (a) The initial state from ERM, as indicated in the left image, shows the target data points are not only distant from the source but also exhibit considerable class overlap, especially within the central region enclosed by the ellipse. 
    This misalignment reflects an accuracy of $63.5\%$, an OT distance of $29.1$, and a prediction entropy of $0.54$. 
    (b) After employing OT-VP, the right image shows that the target representations become more distinct and well-separated, with classes from source and target better aligned.  
    The target data have shifted closer to the corresponding source representations, improving accuracy to $81.4\%$—an increase of \textbf{17.9\%}, and reducing the OT distance and prediction entropy to $25.8$ and $0.27$ respectively. 
    }
    \label{fig:tsne_pacs}
    \vspace{-0.5cm}
\end{figure*}

Remarkably, OT-VP demonstrates significant improvement in the single-source setting. 
In this setting, while methods like T3A and DePT show limited improvements, OT-VP significantly boosts accuracy. 
For example, in the PACS dataset with P as the source and S as the target (P $\rightarrow$ S), T3A increases accuracy by only 1.8\%. 
In contrast, OT-VP improves it by 33.1\%. 
Another case from A to C (A $\rightarrow$C) in PACS sees OT-VP improving accuracy by 17.3\%, against T3A's 5.7\%.
These results are depicted in a t-SNE visualization \ref{fig:tsne_pacs}. For more details, see Table \ref{tab:1src_full_pacs} int the Appendix.

\vspace{12pt}
\noindent\textbf{OT-VP achieves superior performance over training-time prompt learning.} 
Unlike DoPrompt, our approach eliminates the need to delve into the training process, significantly enhancing practical applicability. By leveraging target data, OT-VP adeptly navigates the distributional shifts present in the target domain. It optimizes target-specific prompts to effectively bridge any gaps, moving beyond reliance solely on source domain insights.

DoPrompt, while innovative, learns domain prompts by treating training data as out-of-distribution (OOD) samples. This strategy does not ensure that the prompts will perform well across all potential target domains. Furthermore, DoPrompt requires prompt learning for each domain alongside a prompt adapter (an auxiliary model) to tailor prompts for every target input image. In contrast, our method optimizes universal prompts for a singular target domain, streamlining the process.

As evidenced in Table \ref{tab:1src_avg}, DoPrompt experiences significant performance drops when limited to a single domain, which could limit its applicability in scenarios where source data is scarce, such as in medical imaging. Our method surpasses DoPrompt in both scenarios, underscoring the efficiency and accessibility of learning prompts at test time.

\subsection{Experiments on ImageNet-C}
We conducted further evaluations of our method on ImageNet-C, specifically at the highest severity level of 5. Results are detailed in Table \ref{tab:imagenetc} (top). For DePT, we cite results directly from \cite{gao2023visual}, while outcomes for other methods are derived from three experimental runs. 
DoPrompt was excluded from this analysis because it necessitates changes to the pre-training process, which is incompatible with publicly available pre-trained models. 

\vspace{12pt}
\noindent\textbf{OT-VP boosts accuracy across all corrupted domains.} Although T3A demonstrates performance improvements on PACS, VLCS, and OfficeHome, its effectiveness on ImageNet-C is limited. We hypothesize that the use of 1000 pseudo-prototypes and reliance on distance metrics for predictions complicates adaptation under these conditions.
In contrast, our method, OT-VP, significantly enhances accuracy on ImageNet-C, outperforming T3A across all 15 domains and surpassing DePT in 12 out of 15 domains. Notably, the average accuracy improvement across these domains is 1.9\% over DePT, despite our method only requiring the addition of four prompt tokens to the first transformer layer, whereas DePT integrates 50 prompt tokens across four different layers and updates classifier parameters.

\vspace{12pt}
\noindent\textbf{OT-VP outperforms SOTA in online setting.} We tested our method in an online setting where the model adapts to sequentially incoming target data batches, each used only once for training. We adapted our method by initially training on the first one percent of batches (about 8) for 50 steps and then reducing training to just one step for subsequent batches. This approach assumes all batches originate from the same domain, allowing us to continually refine a universal prompt for all batches, thus speeding up computation. If this domain consistency assumption does not hold, prompts could be updated across all batches for 50 steps to accommodate domain variability. The results, presented at the bottom of Table \ref{tab:imagenetc}, show that OT-VP achieves an average accuracy of 65.7\%, which, while 1.3\% lower than the offline version, is still 2.1\% higher than the online SOTA, DePT, and even slightly surpasses DePT offline by 0.6\%.

In summary, while ERM on ViTs demonstrates significant potential in adapting to unseen target domains, many prior DG and TTA methods struggle to enhance performance \cite{NEURIPS2021_1415fe9f, zheng2022prompt}. Our approach demonstrates considerable improvement across all four datasets, maintaining the entire pre-trained model's parameters while learning only a minimal number of prompt tokens. This showcases our method’s efficiency and effectiveness in enhancing model adaptability without extensive modifications.


\subsection{Ablation Study}
\label{sec:ablation_study}
\textbf{Analysis of the objective functions. } 
To explore the efficacy of our approach, we experimented with directly minimizing entropy during the prompt learning process. 
Interestingly, this direct focus on entropy reduction did not yield improvements. 
Instead, we observed a monotonic decrease in accuracy as entropy was reduced. This suggests that while reducing entropy might intuitively seem beneficial, focusing solely on this aspect can lead to overconfidence in the model's predictions, including those that are incorrect. 
Such over-trust manifests as high confidence in erroneous predictions, underlining a potential pitfall of learning universal visual prompts purely on entropy reduction. 
These findings validate the effectiveness of our chosen objective function, which not only achieves the desired outcome but also implicitly manages entropy without compromising the reliability of predictions (see in App. \ref{sec:entropy_res}).

\begin{figure}[ht]
    \centering
    \begin{subfigure}{0.23\textwidth} 
        \centering
        \includegraphics[width=\linewidth]{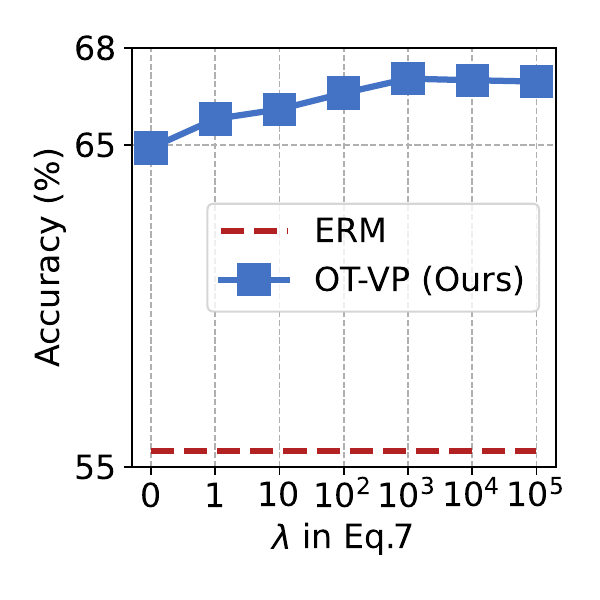} 
        \caption{}
        \label{fig:lambda_ablation}
    \end{subfigure}
    \begin{subfigure}{0.23\textwidth} 
        \centering
        \includegraphics[width=\linewidth]{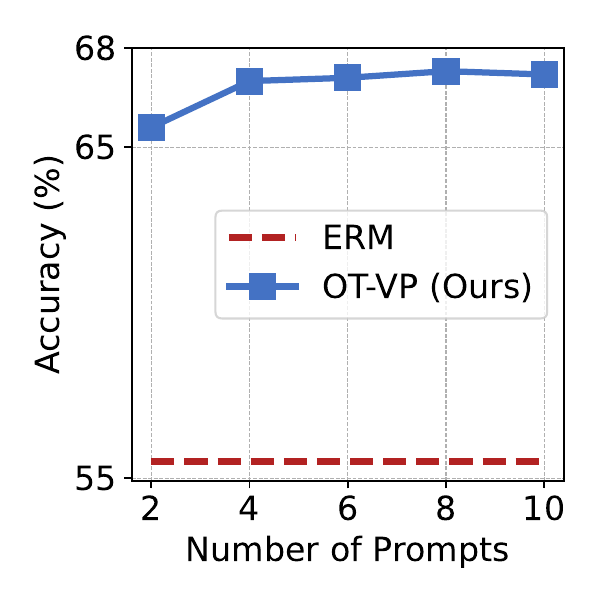} 
        \caption{}
        \label{fig:avg_across_num_prompts}
    \end{subfigure}
    \caption{(a) Influence of hyperparameter $\lambda$ on ImageNet-C. Accuracy declines with smaller $\lambda$ values but stabilizes when $\lambda$ is large. These trends reveal the pivotal role of $\lambda$ in preventing cross-class transport and its impact on overfitting, particularly when using pseudo labels during prompt optimization. (b) Effect of Prompt Length on OT-VP Performance. The performance of OT-VP exhibits only minor variations across different numbers of prompts, showing a robust improvement.}
    \label{fig:images}
\end{figure}

\vspace{12pt}
\noindent\textbf{Analysis on $\lambda$. } In our further investigations, we delved into the impact of the hyperparameter $\lambda$ within Eq. \ref{eq:ot_dist_with_label} on ImageNet-C. 
Our experiments spanned a range of $\lambda$ values: $[0, 1, 10, 10^2, 10^3, 10^4, 10^5]$, where $\lambda=0$ and $\lambda=10^4$ corresponds to the OT-VP-B and OT-VP, respectively. 

Our results highlight two crucial insights: Firstly, incorporating label information into the learned prompts significantly enhances the alignment between the source and target domains, corroborating findings from prior research \cite{alvarez2020geometric}. This label augmentation facilitates a deeper, more meaningful adaptation process. Secondly, we observed a performance decrement with the reduction of $\lambda$ values, as illustrated in Fig. \ref{fig:lambda_ablation}. However, performance reaches a plateau and stabilizes when $\lambda$ is sufficiently large, approaching an effective infinity in relation to feature distance, indicating an optimal balance in the adaptation mechanism. $\lambda = 10^4$ is used for all main experiments.


\vspace{12pt}
\noindent\textbf{Analysis on prompt length.} 
We selected a prompt length of four, in alignment with DoPrompt's configuration. Our experiments, which explored various prompt lengths—2, 4, 6, 8, and 10—demonstrated that the performance of the OT-VP varies only slightly with different numbers of prompts $l$ except for $l=2$, as illustrated in Fig. \ref{fig:lambda_ablation}. This exception likely arises because two prompt tokens are insufficient for adequately capturing domain differences. Consequently, we standardized the prompt length at $l=4$ for all primary experiments. This decision ensures that all parameters are predetermined prior to any interaction with the target data, thereby facilitating a uniform approach across our experimental processes.

\begin{table}[ht]

\scriptsize
\centering
\resizebox{0.5\textwidth}{!}{%
\begin{tabular}{cccccc}
\toprule
\textbf{Algo.} & \textbf{\#Parameters} & \textbf{\#Forwards} & \textbf{\#Backwards} & \textbf{Relative Time} & \textbf{Acc.}  \\ \midrule
ERM & 0 & 782 & 0 & 0.6  & 55.5 \\ 

Tent-BN & 38,400 & 782 & 782 & 1.0  & 59.2 \\ 

T3A & 0 & 782 & 0 & 1.1 & 56.9  \\ 
DePT & 807,168 & 2,346 & 400 & 6.2 &65.1 \\ 
\midrule
\rowcolor{lightgray} 
OT-VP-B & 3,072 & 782 & 100 & 1.0 & 64.9 \\ 
\rowcolor{lightgray} 
OT-VP & 3,072 & 782 & 100 & 1.1 & 67.0 \\ 

\bottomrule
\end{tabular}
}
\caption{Comparative analysis of trainable parameters and computational complexity on ImageNet-C in the Offline Setting. Note that T3A is an optimization-free method requiring prototype updates and generally exhibits suboptimal performance. Our method demonstrates significant improvements. It achieves these advancements with substantially fewer trainable parameters and reduced need for backward propagations than DePT.
}
\label{tab:computation_comparison}

\end{table}
\vspace{12pt}
\noindent\textbf{Analysis of Trainable Parameters and Computational Complexity.} Despite using SGD for optimizing prompt tokens, our OT-VP method remains computationally and memory-efficient. We detail the number of trainable parameters and computation times for all algorithms in Table \ref{tab:computation_comparison}. The trainable parameters for OT-VP consist of only four prompt tokens. In contrast, DePT requires not only a greater number of prompt tokens (50) but also updates to the classifier parameters. Our method outperforms DePT by 1.9\% while utilizing only 0.4\% of the trainable parameters compared to DePT. Furthermore, DePT necessitates additional forward passes due to the processing of three augmented images by both the teacher and student models. In contrast, our method requires only a single forward pass for each image. Additionally, the reduced number of trainable parameters allows our method to stabilize with fewer backward propagations. Consequently, OT-VP can be efficiently applied in real-world scenarios, both in terms of computation and memory usage.

\section{Conclusion}
In this paper, we introduce OT-VP, a novel test-time adaptation approach that utilizes visual prompt learning for effective test-time adaptation. Distinctively, OT-VP adapts without modifying the pre-trained model and effectively aligns the source and target domains, thus providing a more practical alternative to conventional methods. During adaptation, the process requires only the addition of four prompt tokens to the input token sequence, which are then updated using the optimal transport distance between the target representation and a precomputed source representation. We conduct extensive evaluations of the proposed method on both stylistic datasets (PACS, VLCS, OfficeHome) and corrupted datasets (ImageNet-C), demonstrating that learning merely four prompts can achieve state-of-the-art performance across multiple settings. Additionally, we confirm that OT-VP is both memory- and computation-efficient.

{\small
\bibliographystyle{ieee_fullname}
\bibliography{egbib}
}
\appendix
\clearpage

\section{Limitations}
One limitation of our OT-VP approach is its reliance on the quality of pseudo labels for computing the Optimal Transport distance. As visualized in our t-SNE plots \ref{fig:tsne_pacs}, there's a risk of occasional misalignment due to inaccurate pseudo-labeling, which can adversely affect the model's ability to accurately bridge the source and target domain gap. While implementing entropy-based filtering akin to T3A \cite{NEURIPS2021_1415fe9f} could mitigate this by filtering out high-entropy, less reliable pseudo labels, the fundamental limitation remains: OT-VP's capacity to perform effective test-time adaptation may be significantly hindered if the pseudo labels are entirely unreliable or carry no meaningful information about the true class distribution.

\section{Full Results}
\label{sec:full_results}
\subsection{}
In this section, we present the computation time for OT-VP on the PACS dataset. 
We optimize only 4 prompt tokens over 5 epochs, utilizing a 20\% hold-out split from both source and target data. 
While the multi-source setting involves processing triple the data to compute source representations compared to the single-source setting, the time required for both is nearly identical. 
Specifically, the average time is 39.5 seconds for the multi-source and 38.7 seconds for the single-source setting on the PACS dataset on our hardware. 
Moreover, the computational time is slightly influenced by the size of the datasets but remains relatively quick. 
For instance, in the PACS dataset, domain S, which has more than double the data of domain P, requires more processing time—51.7 seconds for S versus 34.5 seconds for P in the multi-source setting. 
Full results for PACS can be found in Table \ref{tab:computation_time_pacs} in Appendix.
In conclusion, OT-VP can efficiently learn prompts in both single and multi-source settings without significant computational overhead.
For the single-source setting, the average computation time is calculated across three different sources.
\begin{table}[h]
    \centering
    \begin{tabular}{lccccc}
        \toprule
        \textbf{Setting} & \textbf{A} & \textbf{C} & \textbf{P} & \textbf{S} & \textbf{Avg} \\
        \midrule
        Single-Source & 32.6 & 37.5 & 33.7 & 50.8 & 38.7 \\
        Multi-Source & 33.9 & 38.0 & 34.5 & 51.7 & 39.5 \\
        \bottomrule
    \end{tabular}
    \caption{Average computation time (seconds) for OT-VP on PACS dataset.}
    \label{tab:computation_time_pacs}
\end{table}

\subsection{}
\label{sec:dataset}
In this section, we provide details for the three stylistic datasets and one corrupted dataset. \textbf{PACS} \cite{li2017deeper} is composed of four domains: \underline{\textbf{P}}hotos, \underline{\textbf{A}}rt, \underline{\textbf{C}}artoon, and \underline{\textbf{S}}ketch, containing 9,991 images in 7 classes.  
\textbf{VLCS} \cite{fang2013unbiased} comprises four real-world photographic datasets: \underline{\textbf{V}}OC2007, \underline{\textbf{L}}abelMe, \underline{\textbf{C}}altech, and \underline{\textbf{S}}UN09, containing 10,729 images in 5 classes.
\textbf{OfficeHome} \cite{Venkateswara_2017_CVPR} consists of four domains: \underline{\textbf{A}}rt, \underline{\textbf{C}}lipart, \underline{\textbf{P}}roduct, \underline{\textbf{R}}eal, containing 15,588 images in 65 classes. \textbf{ImageNet-C} comprises corrupted images in 15 types of corruption. We use the highest level of corruption (\ie severity 5).

\subsection{}
\label{sec:baseline_hyperparameter}
We present the implementation details of our experiments. 
Following \cite{gulrajani2020search}, we partition the data from each domain into training and validation splits of $80\%$ and $20\%$, respectively, utilizing the larger split for training and the smaller one for model selection. 
Our training approach for ERM adheres to the hyperparameters specified by \cite{zheng2022prompt}, incorporating a dropout rate of $0.1$ and a weight decay of $10^{-2}$. 
The learning rate is $5\times 10^{-6}$ for PACS and VLCS, and $10^{-5}$ for OfficeHome. 

For all baseline methods except for DePT, we use their official implementation\footnote{\href{https://github.com/DequanWang/tent}{https://github.com/DequanWang/tent}} \footnote{\href{https://github.com/matsuolab/T3A}{https://github.com/matsuolab/T3A}} \footnote{\href{https://github.com/zhengzangw/DoPrompt}{https://github.com/zhengzangw/DoPrompt}}.
For the implementation of DoPrompt, we set the prompt length to 4, with the coefficient $\lambda$ explored over the set $\{0.1, 1, 10\}$. 
The $M$ parameter for T3A is chosen from $\{1, 5, 20, 50, 100, \mathrm{N/A}\}$, while the configuration for Tent is determined from combinations of $\{0.1, 1.0, 10.0\}$ and $\{1, 3\}$. For DePT, we implement DePT-Group with $M=4$ stages and 50 prompts, adhering to the same hyperparameters specified for ImageNet-C in \cite{gao2023visual}. 

In the single-source scenario, the model is trained on one domain and then adapted to another.
The average accuracy is calculated across all 12 domain pairings for each trial.
In the multi-source setting, one domain is designated as the target while the remaining three serve as sources. 

\subsection{}
In this section, we present the comprehensive outcomes in Tables \ref{tab:1src_avg} and \ref{tab:3src_avg}. The experiments were conducted using three different seeds \{0, 1, 2\} within the \verb|DomainBed| framework. 
Tables \ref{tab:1src_full_pacs}, \ref{tab:1src_full_vlcs}, and \ref{tab:1src_full_officehome} display average results from three rounds for each source-target pair on PACS, VLCS, and OfficeHome, respectively, in the single-source setting. Similarly, Tables \ref{tab:3src_full_pacs}, \ref{tab:3src_full_vlcs}, and \ref{tab:3src_full_officehome} show the average results across three rounds for each target domain on PACS, VLCS, and OfficeHome in the multi-source setting.

\begin{figure}
    \centering
    \includegraphics[width=1\linewidth]{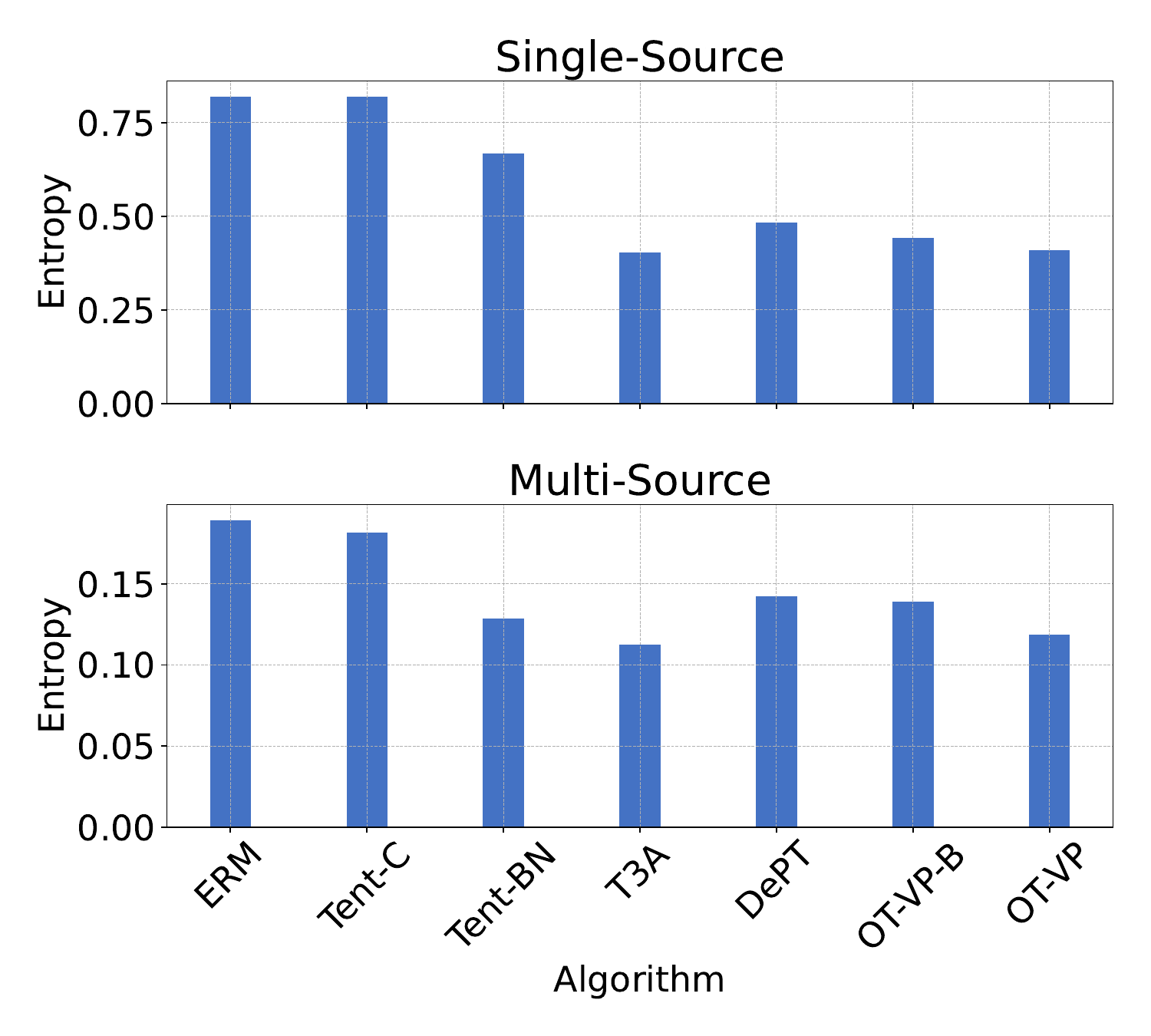}
    \caption{{Prediction entropy across TTA Algorithms in Single-Source and Multi-Source settings on PACS. }
    In both settings, OT-VP demonstrates a marked reduction in entropy, outperforming Tent-C and Tent-BN, which target entropy minimization directly.
    }
    \label{fig:avg_ent_pacs}
    \vspace{-0.5cm}
\end{figure}

\subsection{}
\label{sec:entropy_res}
\vspace{12pt}
\noindent\textbf{OT-VP implicitly reduces prediction entropy.} Consistent with prior research \cite{wang2021tent}, there's an observed correlation between prediction entropy and accuracy—lower entropy often signifies more accurate and confident predictions. 
Unlike traditional approaches that explicitly target entropy reduction by adjusting model parameters \cite{wang2021tent, samadh2023align}, OT-VP achieves this indirectly through the strategic application of Optimal Transport. This involves leveraging a cost metric that encompasses both features and labels \ref{eq:ot_dist_with_label}, aiming to align the target distribution more closely with the source distribution, thereby enhancing model confidence near the decision boundary. 
This alignment is visually supported by representations such as those depicted in Fig. \ref{fig:tsne_pacs}, a t-SNE visualization for source A and the target C (A $\rightarrow$ C) within the PACS dataset.

A comparative analysis of prediction entropy among ERM, Tent-C, Tent-BN, and OT-VP—illustrated in Fig. \ref{fig:avg_ent_pacs}—demonstrates that OT-VP can significantly lower entropy through the refined optimization of prompts. 
Remarkably, it does so even when compared with methods like Tent-C and Tent-BN, which pursue entropy minimization directly. 
It's important to note that the improvements achieved by Tent-C and Tent-BN result from carefully balancing accuracy and entropy reduction when selecting their hyperparameters.

\begin{table}[ht]

\centering

\begin{tabular}{lccccc}

\toprule

\textbf{Algo.} &  & \textbf{A} & \textbf{C} & \textbf{P} & \textbf{S}  \\ 

\midrule

\multirow{4}{*}{\textbf{ERM}} & \textbf{A} & - & 64.5 & 98.9 & 56.4 \\

& \textbf{C} & 83.9 & - & 89.6 & 69.2 \\

& \textbf{P} & 74.2 & 44.4 & - & 34.1 \\

& \textbf{S} & 50.8 & 58.4 & 49.5 & - \\

\midrule

\multirow{4}{*}{\textbf{DoPrompt}} & \textbf{A} & - & 64.6 & 98.5 & 56.5 \\

& \textbf{C} & 84.1 & - & 90.1 & 74.0 \\

& \textbf{P} & 75.6 & 46.2 & - & 35.2 \\

& \textbf{S} & 46.4 & 55.0 & 45.1 & - \\

\midrule

\multirow{4}{*}{\textbf{Tent-C}} & \textbf{A} & - & 64.6 & 98.9 & 56.3 \\

& \textbf{C} & 83.9 & - & 89.6 & 69.0 \\

& \textbf{P} & 74.4 & 44.5 & - & 33.7 \\

& \textbf{S} & 50.4 & 58.3 & 49.0 & - \\

\midrule

\multirow{4}{*}{\textbf{Tent-BN}} & \textbf{A} & - & 71.9 & 98.9 & 66.6 \\

& \textbf{C} & 84.9 & - & 91.3 & 71.7 \\

& \textbf{P} & 78.0 & 56.0 & - & 41.8 \\

& \textbf{S} & 56.3 & 62.9 & 47.5 & - \\

\midrule

\multirow{4}{*}{\textbf{T3A}} & \textbf{A} & - & 70.2 & 98.6 & 67.9 \\

& \textbf{C} & 86.3 & - & 94.4 & 71.1 \\

& \textbf{P} & 80.2 & 53.9 & - & 35.9 \\

& \textbf{S} & 69.0 & 69.9 & 56.9 & - \\

\midrule

\multirow{4}{*}{\textbf{DePT}} & \textbf{A} & - & 70.1 & 98.4& 64.4 \\

& \textbf{C} & 83.9 & - & 90.1 & 69.3 \\

& \textbf{P} & 76.6 & 49.7 & - & 36.1 \\

& \textbf{S} & 51.4 & 63.4 & 52.3 & - \\

\midrule
\midrule

\multirow{4}{*}{\textbf{OT-VP-B}} & \textbf{A} & - & 76.7 & 98.3 & 66.8 \\

& \textbf{C} & 84.4 & - & 92.2 & 69.8 \\

& \textbf{P} & 77.8 & 56.8 & - & 63.9 \\

& \textbf{S} & 44.7 & 58.1 & 40.4 & - \\

\midrule

\multirow{4}{*}{\textbf{OT-VP}} & \textbf{A} & - & 81.8 & 99.0 & 72.2 \\

& \textbf{C} & 84.4 & - & 92.6 & 69.5 \\

& \textbf{P} & 80.4 & 64.3 & - & 67.2 \\

& \textbf{S} & 56.0 & 64.6 & 50.5 & - \\

\bottomrule

\end{tabular}

\caption{Single-Source Full Results on PACS in Table \ref{tab:1src_avg}}

\label{tab:1src_full_pacs}

\end{table}

\begin{table}[ht]

\centering


\begin{tabular}{lccccc}

\toprule

\textbf{Algo.} & \textbf{A} & \textbf{C} & \textbf{P} & \textbf{S} & \textbf{Gain}  \\ 

\midrule

\textbf{ERM} & 91.3 & 82.3 & 98.9 & 75.6 & 87.0 \\

\textbf{DoPrompt} & 91.4 & 81.8 & \textbf{99.5} & 77.1 & \underline{87.5} \\

\textbf{Tent-C} & \underline{91.6} & \underline{82.7} & 98.9 & 75.7 & 87.2 \\

\textbf{Tent-BN} & 91.1 & 82.4 & 98.3 & 76.8 & 87.2 \\

\textbf{T3A} & 91.5 & 81.8 & 99.0 & \underline{77.4} & 87.4 \\

\textbf{DePT} & 91.1 & 81.7 & 99.2 & 77.3 & 87.3 \\

\midrule

\textbf{OT-VP-B} & 91.2 & 81.8 & \underline{99.4} & \textbf{77.4} & 87.3 \\

\textbf{OT-VP} & \textbf{92.0} & \textbf{83.0} & 99.2 & 76.4 & \textbf{87.7} \\

\bottomrule

\end{tabular}


\caption{Multi-Source Full Results on PACS in Table \ref{tab:3src_avg}}

\label{tab:3src_full_pacs}

\end{table}

\begin{table}[ht]

\centering

\begin{tabular}{lccccc}

\toprule

\textbf{Algo.} &  & \textbf{C} & \textbf{L} & \textbf{S} & \textbf{V}  \\ 

\midrule

\multirow{4}{*}{\textbf{ERM}} & \textbf{C} & - & 50.7 & 47.9 & 47.0 \\

& \textbf{L} & 62.9 & - & 55.8 & 63.1 \\

& \textbf{S} & 67.5 & 59.9 & - & 67.7 \\

& \textbf{V} & 96.5 & 66.1 & 80.3 & - \\

\midrule

\multirow{4}{*}{\textbf{DoPrompt}} & \textbf{C} & - & 53.4 & 50.0 & 50.5 \\

& \textbf{L} & 71.7 & - & 57.8 & 70.1 \\

& \textbf{S} & 67.8 & 62.5 & - & 66.2 \\

& \textbf{V} & 98.6 & 62.0 & 78.8 & - \\

\midrule

\multirow{4}{*}{\textbf{Tent-C}} & \textbf{C} & - & 50.4 & 48.3 & 47.0 \\

& \textbf{L} & 70.3 & - & 55.8 & 63.2 \\

& \textbf{S} & 67.2 & 59.8 & - & 67.9 \\

& \textbf{V} & 96.5 & 66.0 & 88.2 & - \\

\midrule

\multirow{4}{*}{\textbf{Tent-BN}} & \textbf{C} & - & 38.3 & 46.9 & 52.7 \\

& \textbf{L} & 50.0 & - & 42.7 & 49.9 \\

& \textbf{S} & 60.9 & 62.3 & - & 69.3 \\

& \textbf{V} & 85.9 & 66.0 & 77.5 & - \\

\midrule

\multirow{4}{*}{\textbf{T3A}} & \textbf{C} & - & 51.8 & 52.1 & 54.3 \\

& \textbf{L} & 83.6 & - & 62.7 & 64.3 \\

& \textbf{S} & 71.1 & 60.5 & - & 67.4 \\

& \textbf{V} & 97.3 & 66.8 & 80.3 & - \\

\midrule

\multirow{4}{*}{\textbf{DePT}} & \textbf{C} & - & 54.6 & 50.8 & 48.5 \\

& \textbf{L} & 78.4 & - & 56.4 & 64.1 \\

& \textbf{S} & 68.4 & 61.2 & - & 67.6 \\

& \textbf{V} & 96.7 & 67.1 & 80.2 & - \\

\midrule
\midrule

\multirow{4}{*}{\textbf{OT-VP-B}} & \textbf{C} & - & 55.7 & 50.0 & 47.0 \\

& \textbf{L} & 73.1 & - & 56.4 & 60.7 \\

& \textbf{S} & 67.1 & 60.8 & - & 67.3 \\

& \textbf{V} & 96.8 & 68.4 & 79.1 & - \\

\midrule

\multirow{4}{*}{\textbf{OT-VP}} & \textbf{C} & - & 59.9 & 51.3 & 48.9 \\

& \textbf{L} & 90.8 & - & 56.3 & 63.8 \\

& \textbf{S} & 69.6 & 64.2 & - & 68.8 \\

& \textbf{V} & 96.8 & 69.3 & 80.8 & - \\

\bottomrule

\end{tabular}

\caption{Single-Source Full Results on VLCS in Table \ref{tab:1src_avg}}

\label{tab:1src_full_vlcs}

\end{table}

\begin{table}[h]

\centering


\begin{tabular}{lccccc}

\toprule

\textbf{Algo.} & \textbf{C} & \textbf{L} & \textbf{S} & \textbf{V} & \textbf{Gain}  \\ 

\midrule

\textbf{ERM} 

& 96.5 & 65.5 & 75.2 & 76.7 & 78.5 \\

\textbf{DoPrompt} & \textbf{98.2} & 67.8 & 75.3 & \textbf{79.9} & \underline{80.3} \\

\textbf{Tent-C} & \underline{97.7} & 65.2 & 75.3 & 76.9 & 78.8 \\

\textbf{Tent-BN} & 86.3 & 66.2 & 68.8 & 72.6 & 73.5 \\

\textbf{T3A} & 97.3 & 65.6 & \textbf{78.0} & \underline{79.3} & 80.0 \\

\textbf{DePT} & 96.6 & 69.2 & 76.7 & 77.8 & 80.1 \\

\midrule

\textbf{OT-VP-B} & 96.8 & \underline{71.9} & 75.2 & 76.9 & 80.2 \\

\textbf{OT-VP} & 96.8 & \textbf{73.1} & \underline{76.8} & 77.0 & \textbf{80.9} \\

\bottomrule

\end{tabular}


\caption{Multi-Source Full Results on VLCS in Table \ref{tab:3src_avg}}

\label{tab:3src_full_vlcs}

\end{table}

\begin{table}[!b]

\centering


\begin{tabular}{lccccc}

\toprule

\textbf{Algo.} &  & \textbf{A} & \textbf{C} & \textbf{P} & \textbf{R}  \\ 

\midrule

\multirow{4}{*}{\textbf{ERM}} & \textbf{A} & - & 54.3 & 71.4 & 77.0 \\

& \textbf{C} & 67.4 & - & 70.0 & 73.2 \\

& \textbf{P} & 62.9 & 47.8 & - & 78.9 \\

& \textbf{R} & 70.3 & 49.2 & 78.5 & - \\

\midrule

\multirow{4}{*}{\textbf{DoPrompt}} & \textbf{A} & - & 52.1 & 71.7 & 79.0 \\

& \textbf{C} & 67.4 & - & 71.7 & 75.5 \\

& \textbf{P} & 66.8 & 47.8 & - & 79.0 \\

& \textbf{R} & 72.2 & 48.8 & 79.6 & - \\

\midrule

\multirow{4}{*}{\textbf{Tent-C}} & \textbf{A} & - & 54.5 & 69.9 & 76.9 \\

& \textbf{C} & 66.9 & - & 69.6 & 73.6 \\

& \textbf{P} & 62.9 & 47.9 & - & 78.6 \\

& \textbf{R} & 71.0 & 47.1 & 79.9 & - \\

\midrule

\multirow{4}{*}{\textbf{Tent-BN}} & \textbf{A} & - & 56.7 & 70.5 & 77.5 \\

& \textbf{C} & 67.9 & - & 70.4 & 72.9 \\

& \textbf{P} & 65.4 & 48.6 & - & 79.1 \\

& \textbf{R} & 72.8 & 49.5 & 79.9 & - \\

\midrule

\multirow{4}{*}{\textbf{T3A}} & \textbf{A} & - & 55.1 & 71.2 & 76.6 \\

& \textbf{C} & 67.9 & - & 71.5 & 74.8 \\

& \textbf{P} & 67.1 & 48.7 & - & 80.3 \\

& \textbf{R} & 72.9 & 49.8 & 80.9 & - \\

\midrule

\multirow{4}{*}{\textbf{DePT}} & \textbf{A} & - & 54.9 & 70.2 & 75.9 \\

& \textbf{C} & 68.1 & - & 70.2 & 73.5 \\

& \textbf{P} & 64.7 & 48.6 & - & 79.2 \\

& \textbf{R} & 71.0 & 49.6 & 79.1 & - \\

\midrule
\midrule

\multirow{4}{*}{\textbf{OT-VP-B}} & \textbf{A} & - & 55.1 & 70.5 & 75.0 \\

& \textbf{C} & 65.8 & - & 69.4 & 73.1 \\

& \textbf{P} & 64.6 & 49.1 & - & 77.4 \\

& \textbf{R} & 71.1 & 52.1 & 79.6 & - \\

\midrule

\multirow{4}{*}{\textbf{OT-VP}} & \textbf{A} & - & 55.0 & 71.4 & 76.9 \\

& \textbf{C} & 67.6 & - & 70.1 & 73.6 \\

& \textbf{P} & 68.7 & 49.7 & - & 79.9 \\

& \textbf{R} & 71.3 & 52.2 & 80.8 & - \\

\bottomrule

\end{tabular}


\caption{Single-Source Full Results on OfficeHome in Table \ref{tab:1src_avg}}

\label{tab:1src_full_officehome}

\end{table}

\begin{table}[b]

\centering


\begin{tabular}{lccccc}

\toprule

\textbf{Algo.} & \textbf{A} & \textbf{C} & \textbf{P} & \textbf{S} & \textbf{Gain}  \\ 

\midrule

\textbf{ERM} & 73.8 & 57.3 & 80.3 & 83.0 & 73.6 \\

\textbf{DoPrompt} & 73.4 & 58.8 & 81.7 & \textbf{84.8} & 74.7 \\

\textbf{Tent-C} & 73.5 & 57.3 & 80.4 & 83.2 & 73.6 \\

\textbf{Tent-BN} & 73.5 & 58.8 & 81.9 & 84.0 & 74.6 \\

\textbf{T3A} & \underline{74.2} & 58.3 & 81.8 & \underline{84.6} & 74.7 \\

\textbf{DePT} & 73.9 & \underline{59.3} & \underline{82.2} & 83.7 & \underline{74.8} \\

\midrule

\textbf{OT-VP-B} & 74.0 & 58.9 & 80.7 & 83.5 & 74.3 \\

\textbf{OT-VP} & \textbf{74.2} & \textbf{59.6} & \textbf{82.3} & 84.1 & \textbf{75.1} \\

\bottomrule

\end{tabular}


\caption{Multi-Source Full Results on OfficeHome in Table \ref{tab:3src_avg}}

\label{tab:3src_full_officehome}

\end{table}

\end{document}